%% file: main.tex
\documentclass[10pt,twocolumn,letterpaper]{article}

\usepackage[T1]{fontenc}
\usepackage[utf8]{inputenc}
\usepackage{authblk}
\usepackage[accsupp]{axessibility}

\usepackage[pagenumbers]{cvpr} 

\usepackage{times}
\usepackage{latexsym}
\usepackage{amsmath}
\usepackage{epsfig}
\usepackage{graphicx}

\usepackage{amssymb}
\usepackage{booktabs}
\usepackage{pifont}

\usepackage{multirow}
\usepackage{bm}
\usepackage{mathtools}
\usepackage{float}
\usepackage{lipsum}

\usepackage{xcolor}
\usepackage{mdframed}
\usepackage{lipsum}
\usepackage{listings}
\usepackage{longtable}
\usepackage[hang,flushmargin]{footmisc}


\definecolor{cvprblue}{rgb}{0.21,0.49,0.74}
\usepackage[pagebackref,breaklinks,colorlinks,allcolors=cvprblue]{hyperref}

\def\paperID{*****} 
\def\confName{CVPR}
\def\confYear{2025}

\title{VAST 1.0: A Unified Framework for \\Controllable and Consistent Video Generation }

\author[1]{Chi Zhang}
\author[1]{Yuanzhi Liang}
\author[1]{Xi Qiu}
\author[1]{Fangqiu Yi}
\author[1]{Xuelong Li\thanks{Corresponding author.}}
\affil[1]{Institute of Artificial Intelligence, China Telecom (TeleAI)}
\affil[ ]{\tt\small zhangc120@chinatelecom.cn, liangyz12@chinatelecom.cn, qiux1@chinatelecom.cn, yifq1@chinatelecom.cn, xuelong\_li@chinatelecom.cn}

\begin{document}

\twocolumn[{%
\maketitle
\begin{figure}[H]
\hsize=\textwidth 
\centering
\includegraphics[width=\textwidth]{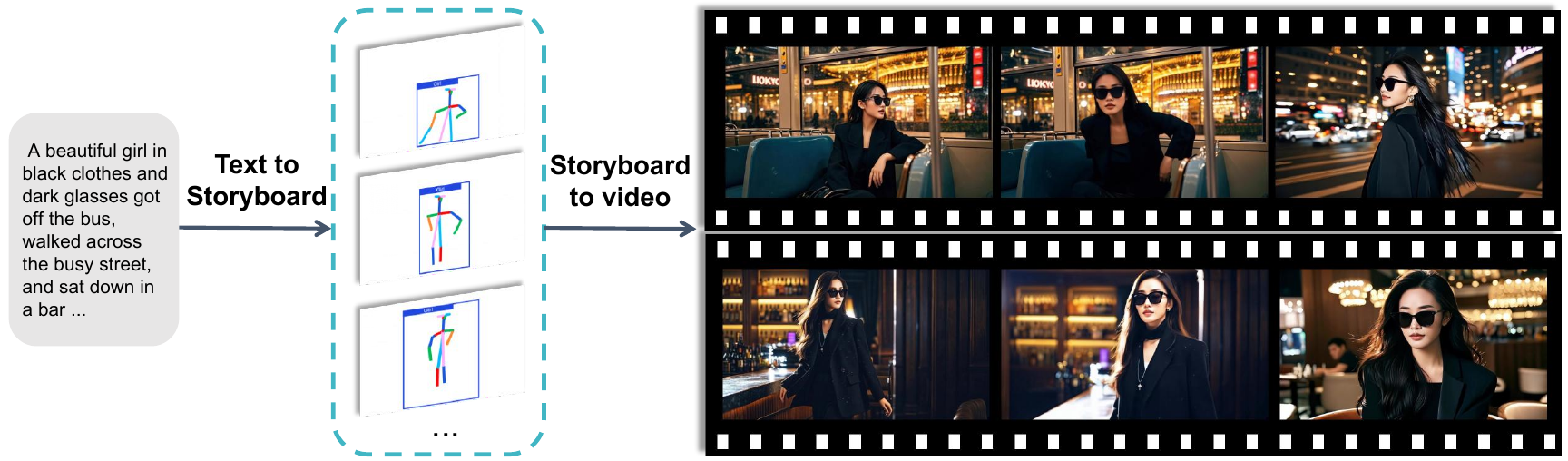}
\caption{Overview of the VAST Framework. The storyboard serves as an intermediary representation, providing precise control signals that enhance temporal consistency, motion dynamics, and spatial coherence in the generated videos. }
\end{figure}
}]

\footnotetext{* Corresponding author. \\ Professor Xuelong Li is the CTO and Chief Scientist of China Telecom, where he founded the Institute of Artificial Intelligence (TeleAI) of China Telecom.}

\input{sec/abstract}    
\input{sec/intro}

\input{sec/related_work}
\input{sec/method}

\input{sec/result}

\input{sec/conclusion}

{
    \small
    \bibliographystyle{ieeenat_fullname}
    \bibliography{main}
}


\end{document}

%% file: sec/abstract.tex
\begin{abstract}

Generating high-quality videos from textual descriptions poses challenges in maintaining temporal coherence and control over subject motion. We propose \textbf{VAST} (Video As Storyboard from Text), a two-stage framework to address these challenges and enable high-quality video generation. In the first stage, StoryForge transforms textual descriptions into detailed storyboards, capturing human poses and object layouts to represent the structural essence of the scene. In the second stage, VisionForge generates videos from these storyboards, producing high-quality videos with smooth motion, temporal consistency, and spatial coherence. By decoupling text understanding from video generation, VAST enables precise control over subject dynamics and scene composition. Experiments on the VBench benchmark demonstrate that VAST outperforms existing methods in both visual quality and semantic expression, setting a new standard for dynamic and coherent video generation.

\end{abstract}

%% file: sec/intro.tex
\section{Introduction}
\label{sec:intro}


Recent advancements in video generation, driven by models such as DiT~\cite{esser2024scaling,peebles2023scalable}, have led to significant improvements in content quality, enabling applications in virtual reality~\cite{poole2022dreamfusion,mo2023dit} and AI model training simulations~\cite{hatamizadeh2025diffit}. Additionally, there has been increasing focus on the development of generative models for constructing world models~\cite{videoworldsimulators2024,zhu2024sora}, recognizing their potential to drive further progress in the field.

However, existing video generation models face limitations, particularly in scenarios requiring complex motion and temporal consistency. Generated videos often lack smooth transitions and fail to represent dynamic motion accurately~\cite{huang2024vbench}, such as running or jumping. Furthermore, these models struggle to follow textual descriptions precisely, leading to unnatural or incomplete motion. For example, when given the prompt “A person diving off a cliff” current models might generate a scene with a person, a cliff, and water, but fail to depict the continuous motion of the dive.

To address these challenges, it is important to consider a more systematic approach to text-to-video generation. Text, as an abstract representation of human semantics, encapsulates a vast amount of information, making it difficult for models to interpret and generate content from this input in a single step. In creative domains such as filmmaking or animation, human creators often begin with storyboards and outlines to organize complex ideas and guide content creation. This structured approach helps break down the task into manageable parts, making it easier to generate coherent and detailed content.

In this work, we introduce storyboarding as a key component of video generation. We believe that generating content directly from text without a clear intermediary structure can complicate the process and make it more challenging for models to produce high-quality results. By generating a storyboard before the video, we simplify the task for the model, allowing it to focus on interpreting the core elements of the scene and generating the corresponding content more effectively. The storyboard provides a useful framework for guiding the video generation process, offering both spatial and semantic control signals that direct the model in a more structured manner.



To this end, we introduce \textbf{VAST} (Video As Storyboard from Text), a two-stage framework designed to address these challenges and generate high-quality, temporally consistent videos. VAST introduces a new storyboard representation that captures both spatial and temporal control signals. This approach decouples text understanding from video generation, using the storyboard to guide the generation process, which helps improve motion dynamics, spatial alignment, and temporal consistency.

VAST consists of two stages. First, we introduce StoryForge, a model that converts textual descriptions into detailed storyboards, including human poses and object layouts. The second stage, VisionForge, generates high-quality videos based on these storyboards and textual prompts. VisionForge leverages diffusion models and learns from comprehensive storyboard representations to produce realistic video sequences with consistent motion.

For training, we curated a large dataset of 100 million images and 30 million high-quality video clips, annotated with captions, object layouts, and human poses. The training process includes three main steps:
1). Training StoryForge using video ground truth data,
2). Training VisionForge with both video and image ground truth data,
3). Jointly training StoryForge and VisionForge using both generated and ground-truth storyboards and videos.

Experimental results reveal that VAST achieves substantial advancements in dynamic control and video generation quality. In the VBench evaluation~\cite{huang2024vbench}, VAST consistently outperforms competitors, securing the top position in 9 out of 16 evaluated categories. These include critical aspects such as temporal stability, object classification accuracy, human action representation, spatial relationship coherence, and overall scene realism. The framework excels particularly in mitigating temporal flickering, achieving near-perfect or perfect scores in multiple metrics—including object classification and human movement, which scored 100\%. Additionally, VAST’s overall semantic expression score reaches an impressive 92.63\%, surpassing the next-best method by 11 points. This significant margin underscores the framework’s capability in ensuring spatial control and semantic consistency. Moreover, VAST achieves a top-tier video quality score of 88.60\%, further demonstrating its ability to produce visually compelling and coherent outputs. In all, VAST offers a robust solution for dynamic and consistent video generation, addressing critical limitations of current models. By leveraging a novel two-stage framework and introducing storyboard representations, VAST advances the field toward more realistic, semantically accurate, and temporally coherent video synthesis.

%% file: sec/related_work.tex
\section{Related Work}

Recent advancements in long video generation have primarily focused on improving video controllability and consistency. A common challenge is ensuring that generated videos adhere closely to the prompt instructions while maintaining consistent instances across the video~\cite{huang2024vbench}. Several approaches have been proposed to tackle these challenges.

To enhance controllability, approaches like Motion-I2V~\cite{shi2024motion} decouple motion and appearance learning by using a two-stage process. In the first stage, a motion field predictor estimates pixel trajectories, and in the second, motion-augmented temporal attention ensures consistent feature propagation across frames. This method offers greater control by enabling sparse annotations of motion trajectories. MotionBooth~\cite{wu2024motionbooth} enables precise object and camera motion control through fine-tuning a text-to-video model with custom object images. It introduces novel loss functions, including Subject Region Loss for improved object region learning and Video Preservation Loss for video consistency. Inference is controlled using Cross-Attention Map Manipulation for object motion and a Latent Shift Module for camera motion.
SlowFast-VGen~\cite{hong2024slowfast} introduces a dual-speed learning system, combining a slow learning phase for world dynamics and a fast learning phase to store episodic memory. This approach improves temporal consistency across long video sequences, outperforming baselines in action-driven video generation with fewer scene cuts and better consistency.

The key idea for improving consistency in video generation may lie in utilizing larger models and more data to leverage scaling laws. Many works, including those based on the DiT architecture~\cite{peebles2023scalable}, adopt this approach to improve video sequence generation due to DiT’s superior performance in generalizing over large datasets.  A notable contribution is MM-DiT~\cite{esser2024scaling}, which proposes a transformer-based architecture for text-to-image tasks. The model integrates rectified flow training and enhances diffusion training formulations for latent models. MM-DiT performs well at scale, showing competitive results compared to proprietary models and offering promising directions for further refinement. CogVideoX~\cite{yang2024cogvideox} builds on this idea, proposing a large-scale text-to-video generation model using diffusion transformers. The model also uses an expert transformer architecture for deep text-video fusion and a progressive training approach to produce coherent, long-duration videos with significant motion. 

%% file: sec/method.tex
\section{VAST Framework}
\label{sec:formatting}

\subsection{Overview}

We present VAST (Video As Storyboard from Text), a unified framework designed for high-quality long video generation from textual descriptions. The framework consists of two core components: StoryForge and VisionForge. StoryForge transforms textual descriptions into storyboards that capture key control signals, like object layouts and human poses, providing spatial and semantic information for realistic video generation. VisionForge then converts these storyboards into dynamic video sequences, ensuring smooth motion and temporal coherence across frames.

To implement the VAST framework, we curated a large-scale dataset comprising over 30 million high-quality video clips and 100 million high-quality images. The dataset was preprocessed by extracting five frames per second and annotating human poses, bounding boxes, and captions for both video clips and images. This rich dataset serves as the foundation for training both StoryForge and VisionForge, enabling the framework to generate high-quality videos.

\begin{figure}[!t]
	\centering
	\includegraphics[width=0.4\textwidth]{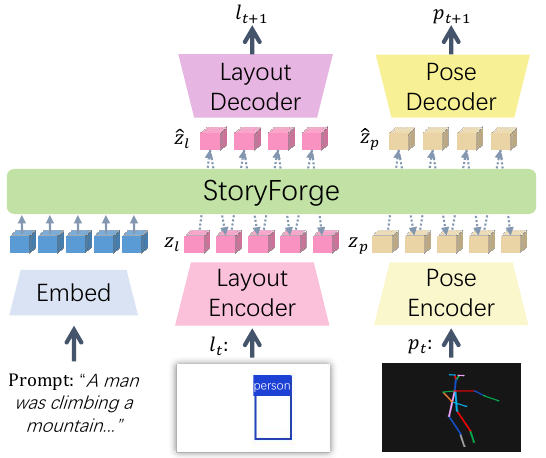}
	\caption{The StoryForge Framework for Text-to-Storyboard Generation. StoryForge converts textual descriptions into detailed storyboards, consisting of human poses, and object layouts. These intermediate representations capture the semantic and structural essence of the input scene, providing precise control signals for subsequent video generation.}
	\label{fig::t2s_framework}
\end{figure}

\subsection{StoryForge: Text-to-Storyboard Generation}

\begin{figure}[!t]
	\centering
	\includegraphics[width=0.5\textwidth]{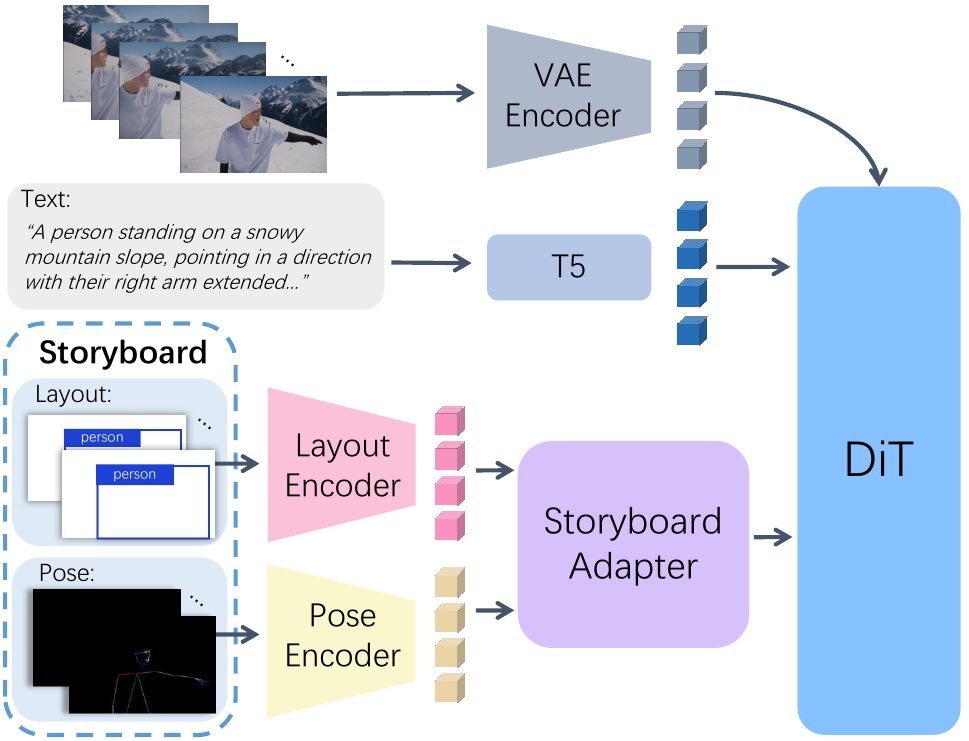}
	\caption{The VisionForge Framework for Storyboard-to-Video Generation. VisionForge takes detailed storyboards as input and synthesizes high-fidelity videos. DiT model ensures dynamic motion, temporal consistency, and spatial coherence by leveraging the structured information encapsulated in the storyboard.}
	\label{fig::s2v_framework}
\end{figure}

StoryForge converts textual descriptions into detailed storyboards for video generation. As shown in Fig.~\ref{fig::s2v_framework}, it consists of a Pose Autoencoder, a Layout Autoencoder, and a Causal Multimodal Large Language Model (MLLM)~\cite{dong2023dreamllm}. The MLLM is introduced to understand both textual semantics and spatiotemporal dependencies from previous time steps, generating corresponding poses and layouts for future steps. This enables StoryForge to generate precise storyboards, guiding the video generation process with both spatial and semantic information.

\textbf{Pose and Layout Autoencoders:} The Pose Autoencoder learns to map pose data to a latent representation. The encoder processes the pose data, and the decoder reconstructs it from the latent space. The reconstruction loss ensures the autoencoder learns accurate pose representations. Similarly, the Layout Autoencoder learns to represent layout data in a latent space. Its encoder processes the layout, and the decoder reconstructs it, with a reconstruction loss function that ensures accurate layout representations.

\textbf{StoryForge Model:} Once the autoencoders are trained, their encoders are used to embed the pose and layout data into latent spaces. These embeddings, along with the input text, are fed into the MLLM. The MLLM learns to generate the next pose and layout features, which are evaluated based on how closely they align with the ground truth. The model is trained to minimize the loss between the generated and the true latent representations of pose and layout.

\subsection{VisionForge: Storyboard-to-Video Generation}

VisionForge is responsible for converting storyboards into dynamic video sequences. As shown in Fig.~\ref{fig::s2v_framework}, it takes human poses, bounding box layouts, and prompt text as input, generating videos that exhibit smooth motion dynamics and temporal coherence. We extend the Diffusion Transformer (DiT)~\cite{esser2024scaling,peebles2023scalable} architecture to handle various input types, including poses, layouts, and text.

The pose and layout encoders from StoryForge are reused to extract features for human poses and object layouts. These features are fused into a unified multimodal representation through the Storyboard Adapter Module, which utilizes self-attention layers to integrate features from the different modalities, making it suitable for video generation tasks. The unified storyboard features are then merged into the DiT model~\cite{polyak2024movie}. 

During training, we employ a multi-phase strategy to optimize VisionForge and StoryForge:

\begin{itemize}
    \item Train StoryForge on the full video dataset to generate detailed storyboards containing poses and layouts.
    \item Train VisionForge on both video and image data, using fixed pose and layout encoders from StoryForge to learn video generation from the storyboard.
    \item Jointly train StoryForge and VisionForge on the full video dataset, randomly selecting storyboard inputs from StoryForge outputs and their corresponding ground-truth annotations to ensure effective collaboration between the models.
\end{itemize}

%% file: sec/result.tex
\section{Experiments}

\subsection{Result in Vbench}

We evaluate our proposed VAST framework on the VBench leaderboard, comparing its performance against state-of-the-art video generation models across multiple metrics. The experimental results demonstrate VAST's strengths in dynamic control, temporal consistency, and semantic accuracy.
\begin{table}[!t]
	\begin{center}
	\resizebox{1.05\columnwidth}{!}
      {
        \begin{tabular}{|l|c|c|c|c|}
        \hline
        Model name       & Rank & Total Score & Quality Score & Semantic Score \\ \hline \hline 
        \textbf{VAST}                             & \textbf{1}    & \textbf{89.71 }      & \textbf{88.98}         & \textbf{92.63}          \\ \hline
        StratoLens                       & 2    & 83.78       & 84.39         & 81.34          \\ \hline
        JT-CV-9B                         & 3    & 83.69       & 84.62         & 79.97          \\ \hline
        T2V-Turbo-v2~\cite{li2024t2v}                     & 4    & 83.52       & 85.13         & 77.12          \\ \hline
        MiniMax-Video-01                 & 5    & 83.41       & 84.85         & 77.65          \\ \hline
        HunyuanVideo~\cite{kong2024hunyuanvideo}                                      & 6     & 83.24       & 85.09         & 75.82          \\ \hline 
        CausVid~\cite{yin2024causvid}                          & 7    & 82.85       & 83.71         & 79.40          \\ \hline
        Data-Juicer~\cite{chen2024data}                      & 8    & 82.53       & 83.38         & 79.13          \\ \hline
        Gen-3                            & 9    & 83.32       & 84.11         & 75.17          \\ \hline
        Vchitect-2.0                     & 10    & 82.24       & 83.54         & 77.06          \\ \hline
        \end{tabular}
     }
     \end{center}
    \caption{Quantitative Comparison of Video Generation Models on VBench. VAST achieves the highest overall score, outperforming other methods in both quality and semantic accuracy.}
    \vspace{-.1in}
    \label{table::tab_overall}
\end{table}
As shown in Tab.~\ref{table::tab_overall}, VAST achieves the highest overall score of 89.71, significantly outperforming competing methods such as StratoLens (83.78) and JT-CV-9B (83.69). VAST also achieves the highest quality score (88.98) and semantic score (92.63), highlighting its ability to produce videos that are not only visually appealing but also semantically coherent. This superior performance reflects the efficacy of our two-stage framework in addressing the challenges of high-dynamic motion and temporal consistency.

To further illustrate the high quality results VAST achieved, we analyze its performance across various fine-grained metrics presented in Tab.~\ref{table::tab_qua}, VAST excels in key aspects such as subject consistency (97.68), background consistency (98.57), and temporal flickering (99.87), indicating its ability to generate stable and coherent videos. Additionally, VAST demonstrates superior motion smoothness (97.99) and dynamic degree (97.22), capturing complex motion sequences more effectively than other models.

In comparison, other methods show inconsistencies in these metrics. For instance, while StratoLens performs well in background consistency (95.57), it falls short in dynamic degree (98.61), indicating limited capability in generating diverse and complex motions. Similarly, MiniMax-Video-01 achieves high temporal flickering performance (99.10) but struggles with dynamic degree (64.91).

\begin{table*}[!t]
	\begin{center}
	\resizebox{1.8\columnwidth}{!}
      {
            \begin{tabular}{|l|c|c|c|c|c|c|c|}
            \hline
            Model name       & \begin{tabular}[c]{@{}c@{}}Subject \\ Consistency\end{tabular} & \begin{tabular}[c]{@{}c@{}}Background\\ Consistency\end{tabular} & \begin{tabular}[c]{@{}c@{}}Temporal\\ Flickering\end{tabular} & \begin{tabular}[c]{@{}c@{}}Motion\\ Smoothness\end{tabular} & \begin{tabular}[c]{@{}c@{}}Dynamic\\ Degree\end{tabular} & \begin{tabular}[c]{@{}c@{}}Aesthetic\\ Quality\end{tabular} & \begin{tabular}[c]{@{}c@{}}Imaging\\ Quality\end{tabular} \\ \hline \hline 
            VAST             & 97.68                                                          & 98.57                                                            & 99.87                                                         & 97.99                                                       & 97.22                                                    & 69.01                                                       & 71.77                                                     \\ \hline
            StratoLens       & 89.20                                                          & 95.57                                                            & 99.08                                                         & 96.64                                                       & 98.61                                                    & 65.60                                                       & 65.44                                                     \\ \hline
            JT-CV-9B         & 94.88                                                          & 95.64                                                            & 97.24                                                         & 98.55                                                       & 95.28                                                    & 62.36                                                       & 63.54                                                     \\ \hline
            T2V-Turbo-v2~\cite{li2024t2v}         & 95.50                                                          & 96.71                                                            & 97.35                                                         & 97.07                                                       & 90.00                                                    & 62.61                                                       & 71.78                                                     \\ \hline
            MiniMax-Video-01 & 97.51                                                          & 97.05                                                            & 99.10                                                         & 99.22                                                       & 64.91                                                    & 63.03                                                       & 67.17                                                     \\ \hline
            HunyuanVideo~\cite{kong2024hunyuanvideo}                       & 97.37               & 97.76             &   99.44            &        98.99                 & 70.83            &    60.36      &   67.56  \\ \hline 
            CausVid~\cite{yin2024causvid}  & 93.33                                                          & 96.65                                                            & 93.85                                                         & 97.64                                                       & 87.22                                                    & 66.49                                                       & 70.20                                                     \\ \hline
            Data-Juicer~\cite{chen2024data}   & 97.92                                                          & 99.27                                                            & 98.14                                                         & 97.77                                                       & 38.89                                                    & 67.39                                                       & 70.41                                                     \\ \hline
            Gen-3            & 97.10                                                          & 96.62                                                            & 98.61                                                         & 99.23                                                       & 60.14                                                    & 63.34                                                       & 66.82                                                     \\ \hline
            Vchitect-2.0     & 96.83                                                          & 96.66                                                            & 98.57                                                         & 98.98                                                       & 63.89                                                    & 60.41                                                       & 65.35                                                     \\ \hline
            \end{tabular}
     }
     \end{center}
    \caption{Evaluation of quality related metrics. VAST excels in key metrics such as subject consistency, background consistency, temporal flickering, and motion smoothness, demonstrating superior temporal coherence and dynamic control.}
    \vspace{-.1in}
    \label{table::tab_qua}
\end{table*}

\begin{table*}[!t]
	\begin{center}
	\resizebox{2\columnwidth}{!}
      {
        \begin{tabular}{|l|c|c|c|c|c|c|c|c|c|}
        \hline
        Model name       & \begin{tabular}[c]{@{}c@{}}Object\\ Class\end{tabular} & \begin{tabular}[c]{@{}c@{}}Multiple\\ Objects\end{tabular} & \begin{tabular}[c]{@{}c@{}}Human\\ Action\end{tabular} & Color & \begin{tabular}[c]{@{}c@{}}Spatial\\ Relationship\end{tabular} & Scene & \begin{tabular}[c]{@{}c@{}}Appearance\\ Style\end{tabular} & \begin{tabular}[c]{@{}c@{}}Temporal\\ Style\end{tabular} & \begin{tabular}[c]{@{}c@{}}Overall\\ Consistency\end{tabular} \\ \hline \hline 
        VAST             & 100                                                    & 99.16                                                      & 100                                                    & 99.76 & 97.02                                                          & 79.22 & 25.98                                                      & 28.04                                                    & 26.72                                                         \\ \hline
        StratoLens       & 78.06                                                  & 71.72                                                      & 100                                                    & 88.41 & 85.93                                                          & 57.08 & 23.73                                                      & 25.92                                                    & 27.40                                                         \\ \hline
        JT-CV-9B         & 91.90                                                  & 69.48                                                      & 95.20                                                  & 89.66 & 75.41                                                          & 57.28 & 23.95                                                      & 25.25                                                    & 27.37                                                         \\ \hline
        T2V-Turbo-v2~\cite{li2024t2v}     & 95.33                                                  & 61.49                                                      & 96.20                                                  & 92.53 & 43.32                                                          & 56.40 & 24.17                                                      & 27.06                                                    & 28.26                                                         \\ \hline
        MiniMax-Video-01 & 87.83              & 76.04            & 92.40           & 90.36 & 75.50           & 50.68 & 20.06                  & 25.63          & 27.10          \\ \hline
        HunyuanVideo~\cite{kong2024hunyuanvideo}                   & 86.10               & 68.55           & 94.40         & 91.60    & 68.68      & 53.88    & 19.80        & 23.89     & 26.44    \\ \hline  
        CausVid~\cite{yin2024causvid}          & 93.70                                                  & 74.68                                                      & 99.80                                                  & 82.37 & 66.72                                                          & 55.86 & 24.13                                                      & 25.23                                                    & 27.52                                                         \\ \hline
        Data-Juicer~\cite{chen2024data} & 96.44                                                  & 64.51                                                      & 95.40                                                  & 95.51 & 47.17                                                          & 57.30 & 25.55                                                      & 26.82                                                    & 29.25                                                         \\ \hline
        Gen-3            & 87.81                                                  & 53.64                                                      & 96.40                                                  & 80.90 & 65.09                                                          & 54.57 & 24.31                                                      & 24.71                                                    & 26.69                                                         \\ \hline
        Vchitect-2.0     & 86.61                                                  & 68.84                                                      & 97.20                                                  & 87.40 & 57.55                                                          & 56.57 & 23.73                                                      & 25.01                                                    & 27.57                                                         \\ \hline
        \end{tabular}
     }
     \end{center}
    \caption{Evaluation of semantic related metrics. VAST leads in object classification, human action, and spatial relationships, reflecting its ability to generate semantically accurate and coherent videos.}
    \vspace{-.1in}
    \label{table::tab_sem}
\end{table*}

\footnotetext{StratoLens: https://stratolens.github.io}
\footnotetext{JT-CV-9B: https://github.com/jiutiancv}
\footnotetext{MiniMax-Video-01: https://platform.minimaxi.com}
\footnotetext{Gen-3: https://runwayml.com/product}
\footnotetext{Vchitect-2.0: https://github.com/Vchitect/Vchitect-2.0}

Tab.~\ref{table::tab_sem} further underscores VAST's strength in semantic understanding and spatial relationships. VAST achieves a perfect score of 100 in object classification and human action, outperforming all other methods. It also excels in maintaining accurate spatial relationships (97.02) and demonstrates outstanding color accuracy (99.76). These results reflect VAST's ability to generate videos that adhere closely to the semantics of the input text and storyboard constraints.

Competing models like StratoLens and JT-CV-9B perform well in certain aspects but lack the comprehensive balance seen in VAST. For example, StratoLens achieves 100 in human action but falls short in object classification (78.06), while JT-CV-9B struggles with multiple objects (69.48).

Overall, VAST demonstrates a clear advantage in producing dynamic, temporally consistent, and semantically accurate videos. Its innovative two-stage approach—leveraging detailed storyboard representations—effectively mitigates the challenges faced by existing models, resulting in superior performance across a wide range of metrics. These experimental results validate VAST as a robust solution for high-fidelity video generation, advancing the field toward more realistic and semantically coherent video synthesis.



\subsection{Results for video generation}

\begin{figure*}[!t]
	\centering
	\includegraphics[width=0.7\textwidth]{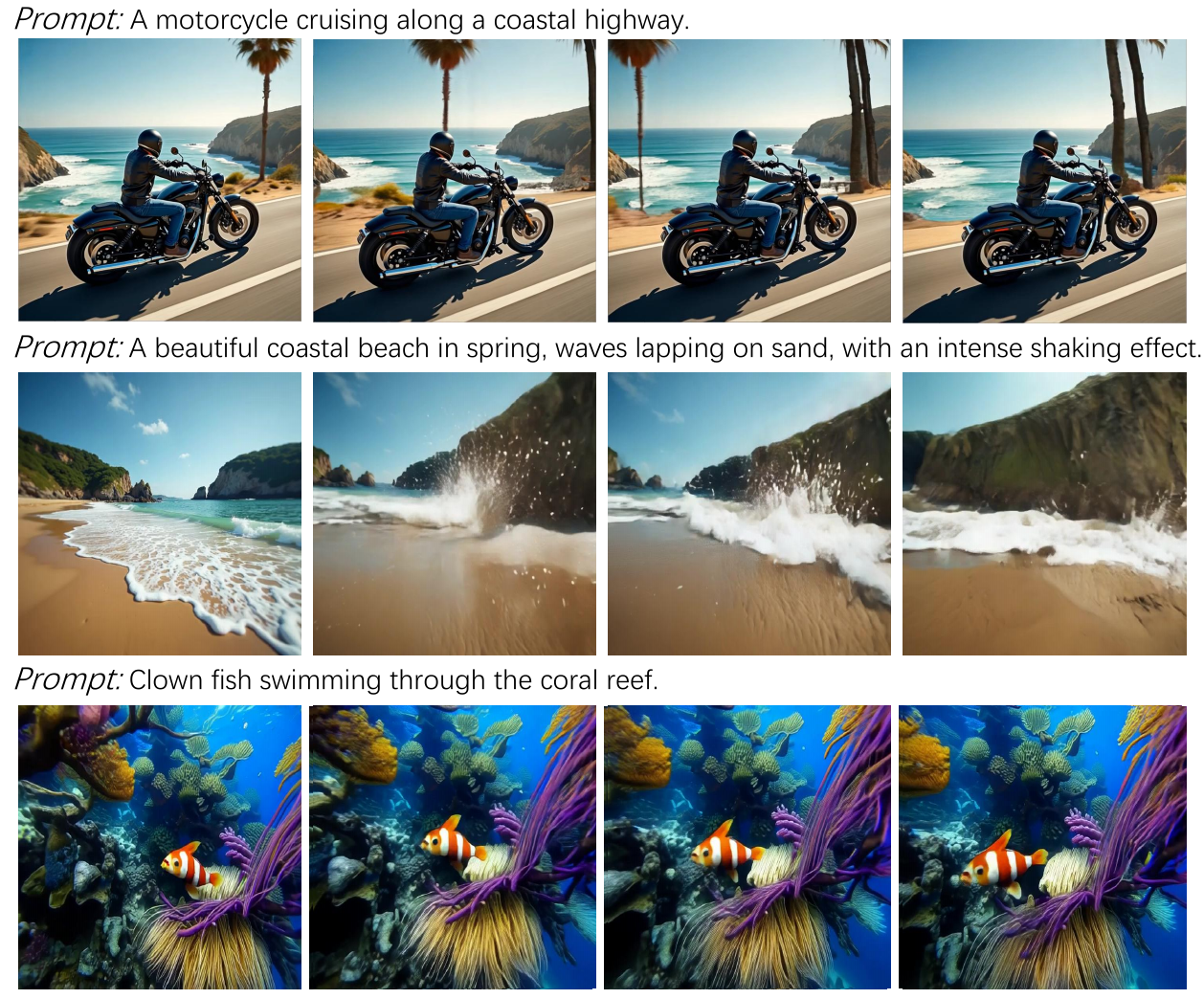}
	\caption{Qualitative Results of VAST on VBench Prompts. Examples of videos generated by VAST in response to VBench prompts, demonstrating superior temporal consistency, dynamic motion, and semantic accuracy.}
	\label{fig::vb_case}
\end{figure*}

To further demonstrate the strengths of VAST, we present qualitative results that showcase the high-fidelity and dynamic videos generated by our method. Figures illustrate VAST's performance using prompts from the VBench benchmark, highlighting its ability to generate coherent and consistent motions across diverse scenarios. For example, when prompted with ``A beautiful coastal beach in spring, waves lapping on sand, with an intense shaking effect'', VAST captures the entire sequence with temporal consistency and realistic motion dynamics.

\begin{figure*}[!t]
	\centering
	\includegraphics[width=\textwidth]{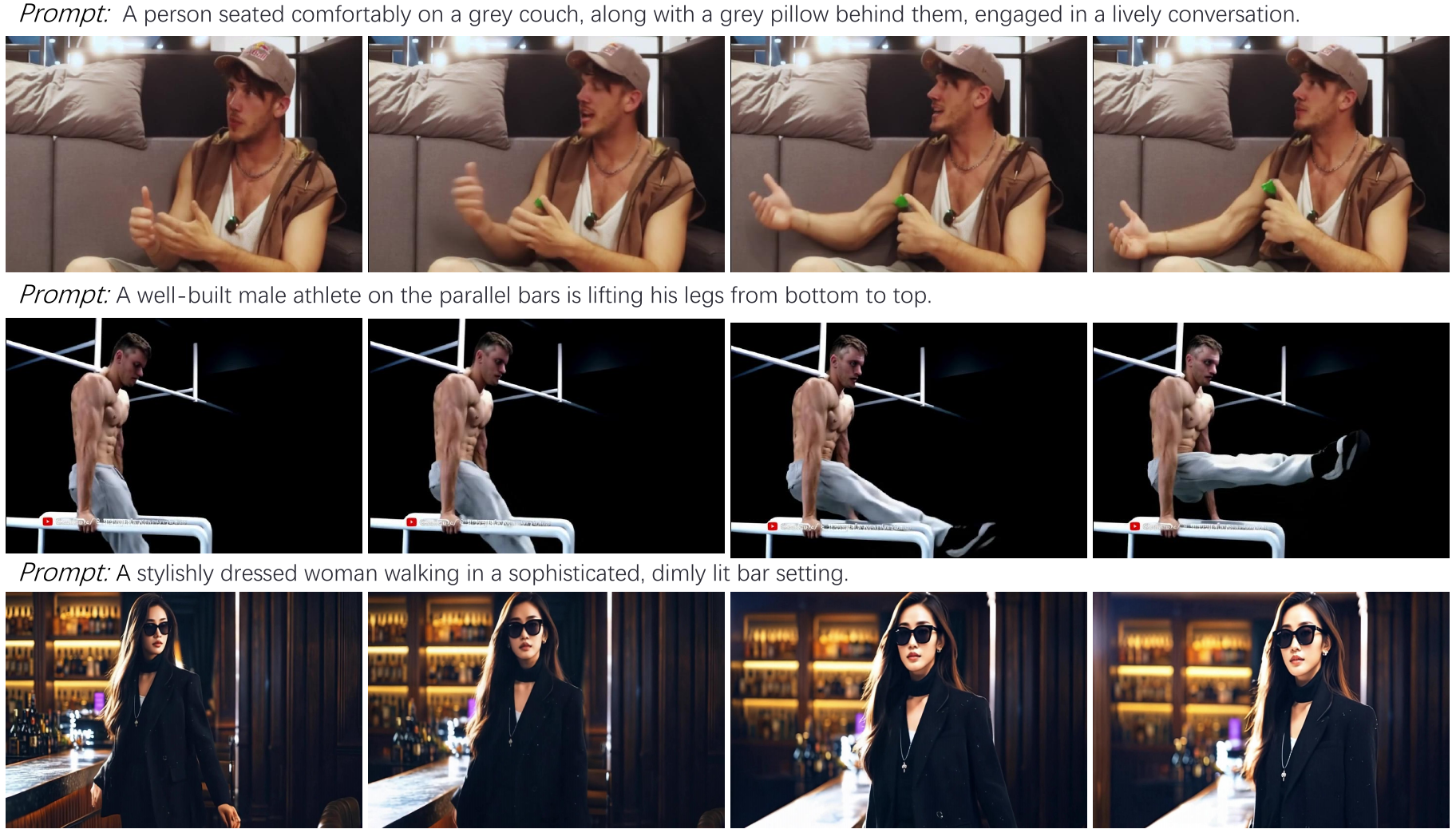}
	\caption{High-Quality Motion Generation by VAST. Videos generated by VAST demonstrating realistic motion dynamics. These examples highlight VAST's ability to intricate action sequences, capturing temporal consistency and dynamic complexity that surpasses existing methods.}
	\label{fig::motion_case}
\end{figure*}

In addition, we present examples of high-quality videos featuring vivid and dynamic motions, such as daily conversations, walking, and even fighting. These animations are characterized by smooth transitions, consistent subject appearances, and realistic interactions with the environment. The results demonstrate that VAST excels not only in quantitative benchmarks but also in generating visually impressive and realistic outputs. These capabilities make VAST suitable for real-world applications such as virtual reality, automated content creation, and simulation environments.

These qualitative evaluations reinforce VAST’s strengths in producing videos that are both dynamic and semantically coherent, effectively addressing the limitations of existing methods and pushing the boundaries of video generation technology.

\begin{figure*}[!t]
	\centering
	\includegraphics[width=\textwidth]{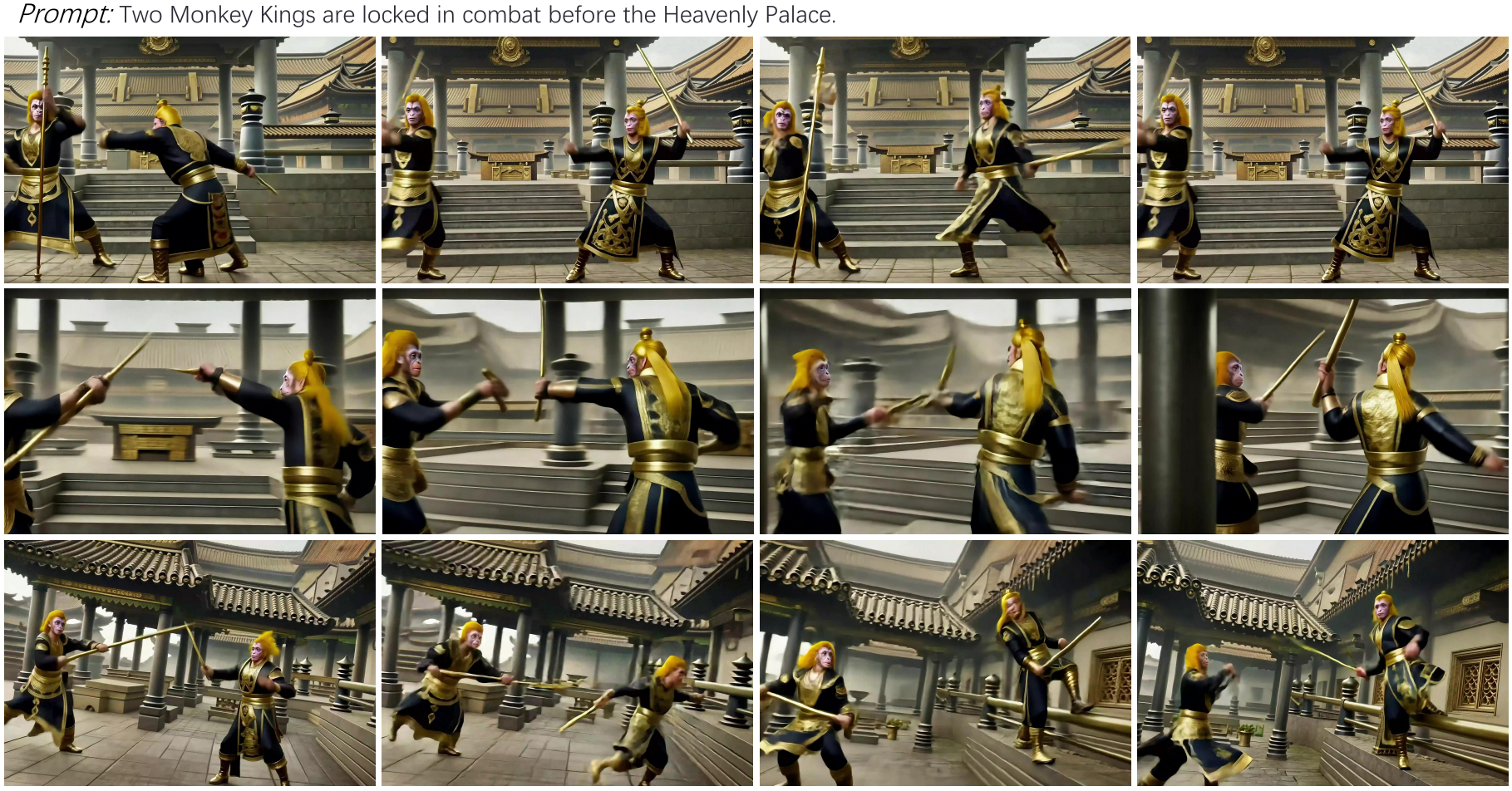}
	\caption{High-Quality Motion Generation by VAST. Videos generated by VAST demonstrating realistic motion dynamics. These examples highlight VAST's ability to intricate action sequences, capturing temporal consistency and dynamic complexity that surpasses existing methods.}
	\label{fig::monkey_case}
\end{figure*}

%% file: sec/conclusion.tex
\section{Conclusion}
We present VAST, a unified framework for high-quality text-to-video generation, consisting of StoryForge and VisionForge. By using intermediate representations, VAST captures both spatial and temporal dynamics, ensuring videos are semantically accurate and temporally coherent.
StoryForge converts text descriptions into detailed storyboards, while VisionForge generates dynamic video sequences from these storyboards. This two-stage approach provides fine-grained control over object movements and visual styles, making the framework adaptable to various scenarios.
VAST combines the power of diffusion models with interpretable intermediate representations, improving spatial alignment and temporal coherence. It sets a new standard for scalable, controllable video generation, addressing key challenges in the field.

\section{Future work}
In future work, we aim to address several limitations of our current approach. One key challenge is the model’s potential difficulty in handling diverse art styles, as it has primarily been trained on real-world data. This focus may lead to insufficient performance when generating scenes with stylistic variations or artistic renderings. Additionally, the generation of objects and animals remains underdeveloped, as our current model may not capture all the intricacies of these elements in certain contexts.

To overcome these issues, we plan to incorporate more diverse and stylized data during training, ensuring the model can handle a broader range of visual styles. Furthermore, integrating 3D information is an important direction for enhancing robustness and consistency, particularly for complex scenes and objects. By incorporating 3D-aware data into the model, we aim to improve its capacity to generate more realistic and varied content across a wider range of scenarios. This would also steer the model toward evolving into a more comprehensive world model, capable of adapting to a diverse array of dynamic environments.